\documentclass{llncs}
\usepackage{llncsdoc}
\usepackage{amsmath,graphicx}
\usepackage{amssymb}
\usepackage{subfigure}
\usepackage{array}
\newcolumntype{P}[1]{>{\centering\arraybackslash}p{#1}}

\usepackage{tabularx}
\usepackage{graphics}

\newcommand{\upperRomannumeral}[1]{\uppercase\expandafter{\romannumeral#1}}

\title{Chest X-rays Classification \\ a Multi-Label and Fine-Grained Problem}
\author{Zongyuan Ge$^{1}$, Dwarikanath Mahapatra$^{2}$, Suman Sedai$^{2}$, Rajib Chakravorty$^{2}$, Rahil Garnavi$^{2}$}

\authorrunning{Ge et. al.}
\institute{$^{1}$ Monash University, Melbourne, Australia\\
  $^{2}$ IBM Research Australia, Melbourne \\ 
\email{zongyuan.ge@monash.edu, \{dwarim,ssedai,rahilgar\}@au1.ibm.com, rajib.chakravorty@gmail.com}}

\begin{document}

\maketitle

\vspace*{-0.4cm}
\begin{abstract}
\label{sec:abstract}

ChestX-ray14 dataset is considered as a general image classification problem in existing literature. This specific dataset has properties-1) many lung pathologies are visually similar, 2) a wealth of diseases including lung cancer, tuberculosis, and pneumonia may present in a single scan (multiple labels) and 3) The number of normal instances is far more than abnormalities (imbalance data). These properties are common in medical domain. Existing literature use state-of-the-art DensetNet/Resnet models being transfer learned where output neurons of the networks are trained for individual diseases to cater for multiple diseases labels in each image. However, in this work we have proposed a novel error function based on softmax concept (Multi-label Softmax Loss, MSML) to address the problems of multiple labels and imbalance data. We have designed convolutional deep network based on fine-grained classification methodology that incorporates MSML. We have evaluated our proposed method on various network backbones and showed performance improvements of AUC-ROC scores on the ChestX-ray14 dataset. The proposed error function provides a new direction to attain improved performance for wider medical data.


%
%
\end{abstract}





\section{Introduction}
\label{sec:introduction}

Chest X-ray is one of the most common radiology exams used for screening of lung diseases.
X-ray is econonimical and can be performed with minimal procedural steps. Moroever, each scan can detect multiple suspected pathologies such as tuberculosis and pneumonia, etc.
Computer-Aided Detection (CADe) and Diagnosis (CADx) has been a major research focus in medicine. CAD for chest X-ray could potentially become a cost-effective assistive tool for radiologists.

Recent advances in artificial intelligence and machine learning have demonstrated that deep learning technologies have superiority in solving various X-ray analysis tasks involving image classification~\cite{wang2017chestx,yao2017learning,rajpurkar2017chexnet}, NLP based analysis~\cite{shin2016learning} and localisation~\cite{payer2016regressing}. 
The data-driven nature of deep learning benefits from the increasing volume of publicly accessible medical imaging dataset, such as CT, MRI, X-ray~\cite{armato2011lung,wang2017chestx}.
Wang et al.~\cite{wang2017chestx} introduced a large scale collection of chest X-ray which contains the absence or presence of 14 lung diseases. 
Several methods explored the use of ResNet~\cite{wang2017chestx} and DenseNet~\cite{rajpurkar2017chexnet,yao2017learning}, that were pre-trained on ImageNet~\cite{russakovsky2015imagenet}, to classify 14 chest related diseases.

However, there are a few challenges which may obstruct the improvement gained from the usage of direct transfer learning from general image classification tasks to detecting lung diseases in chest x-ray and, more generally, in wider medical imaging domain. First unlike categories in ImageNet where many of them are from different branches of WordNet, the high visual similarity among a wealth of pathologies in lung X-rays can be difficult to interpret and distinguish. Second, presence of patterns from various potential pathologies in one medical scan mandates the model learn a huge number of possible label sets for prediction, which is exponential to the size of label space ($2^{C}$ for $C$ labels). Third, issues like the class-imbalance among disease labels and overwhelming normal samples, cannot be fully taken into consideration by just using sigmoid with binary cross-entropy loss~\cite{wang2017chestx,rajpurkar2017chexnet} as in done in standard transfer learning based methodologies. These challenges drive a need for innovative learning mechanisms which consider both subtle differences among various classes, as well as multi-label nature of the task. 

This paper introduces two contributions to detect lung diseases from chest x-ray.

\begin{enumerate}
\item First, we provide a fine-grained perspective into this problem, where fine-grained image classification is defined as a problem to categorise visually similar sub-categories~\cite{farrell2011birdlets}. Thus, this motivates us to explore and re-adapt bilinear pooling method~\cite{lin2015bilinear} from fine-grained classification field. 
\item Second, a neural network loss named Multi-label SoftMax Loss (MSML), is proposed, that captures the characteristics of multi-label learning.  We further combine the idea of MSML with bilinear pooling to propose a novel mecahnism for multi-label learning in a deep learning context.
\end{enumerate}
\section{Methods}
\label{sec:methods}
\vspace*{-0.2cm}
Formally speaking, let $Y=\{1,2,...,C\}$ be the finite set of labels and $Train=\{(input^{n},y^{n})\}_{n=1}^{N}$ be the training set where $input^n$ denotes $n-th$ input and $y^n$ is its relevant label. The label $y \in \{0,1\}^{C}$, where $y_{c}=1$ if the $c$-th label is present and $0$ otherwise. 
$x \in (X=\mathbb{R}^{d})$ denotes the input space of $d$-dimensional feature vectors.
The goal of our problem is to learn a multi-label hypothesis with parameters $\theta$, $h_{\theta}:\mathbb{R}^{d} \rightarrow \mathbb{R}^{C}$ to output predictions to each class accurately. 
 
\subsection{Multi-Label Learning Loss}
The error function of the hypothesis on the training set is defined as $E = \sum_{i=1}^{N}E^{i}$
where $E^{i}$ is the error of the hypothesis on the $i$-th input sample. Using sigmoid as activation function ($sigmoid(.)=\frac{1}{1+\exp(.)}$ in the classification layer, $E_{i}$ could be defined as $E^{i,sigmoid} = -\sum_{c=1}^{C}(y^{i}_{c}log(z^{i}_{c})+(1-y^{i}_{c})log(1-z^{i}_{c})) $ where $z^{i}_{c}$ is the sigmoid output of input feature $x^{i}$ on the $c$-th class, which takes the value between $0$ and $+1$. During training, optimisation method such as stochastic gradient descent (SGD) can be directly applied to learn from the multi-label annotations. 

\noindent Our goal is to minimise the errors from the training set for the model $h_{\theta}$. To look into the gradient being backpropogated, we compute the derivative of the error with respect to each class using the chain rule, $\frac{\partial E}{\partial x_{c}} = \frac{\partial E}{\partial z_{c}} \frac{\partial z_{c}}{\partial x_{c}}$. Following~\cite{rumelhart1985learning}, with $\frac{\partial E^{sigmoid}}{\partial z_{c}}=-\frac{z_{c}-y_{c}}{z_{c}(1-z_{c})}$:
\begin{equation}
\frac{\partial E}{\partial x_{c}} = z_{c} - y_{c} = \frac{1}{1+\exp({-x_{c}})} - y_{c}\label{eq:deledelx}
\end{equation}
However, the error function $E$ in (\ref{eq:deledelx} ) only considers individual class discrimination and does not consider interdependencies among target classes explicitely. Therefore we propose a novel loss function for deep learning based multi-label learning task that considers the relationship of multiple labels explicitely. In this paper, the property of leveraging correlations of multi-label is addressed by the MSML (Multi-label Softmax Loss) error function as follows:
\begin{equation}
E^{MSML} = \frac{1}{|Y_{i}|}\sum_{l \in Y_{i}} \frac{\exp(x_{l}^{i})} { \exp(x_{l}^{i})+\sum_{k \in \bar{Y_{i}} }\exp(x_{k}^{i})}  
\end{equation}
$Y_{i}$ defines the positive class indices $y_{c}=1$ of the current sample, where $\bar{Y_{i}}$ is the complementary of $Y_{i}$ ($y_{c}$=0). $|Y_{i}|$ measures the cardinality and is used for normalisation.
MSML is bootstrapped from softmax function $\frac{exp(x_{c})}{\sum_{C}exp(x_{c})}$.     
Each positive response $x_{l \in Y_{i}}$ is fed to the exponential function. The denominator contains activations from all negative outputs $x_{l \in \bar{Y_{i}}}$ and the positive activation from the nominator. 

We show that MSML is simple to evaluate and differentiate. Using the similar approach as standard error, we can backpropagate the gradient from our proposed MSML. With $\frac{\partial E^{MSML}}{\partial z_{c}}=-\frac{y_{c}}{z_{c}}$, then the overal gradient  is calculated as,
\begin{flalign}
\frac{\partial E^{\text{MSML}}}{\partial x^{i}} & = \sum_{c=1}^{C} \frac{\partial E^{\text{MSML}}}{\partial z_{c}}\frac{\partial z_{c}}{\partial x^{i}} &&\\\nonumber
& = \frac{1}{|Y_{i}|} \sum_{j=1,l_{j} \in Y_{i}}^{Y_{i}} \sum_{c=1}^{C} ( \frac{\partial E^{\text{MSML}}}{\partial z_{c}} \frac{\partial z_{c}}{\partial x_{l_{j}}^{i}} - \sum_{k \neq c, k \in \bar{Y_{i}}} \frac{\partial E^{\text{MSML}}}{\partial z_{c}} \frac{\partial z_{c}}{\partial x_{l_{j}}^{i}}   ) &&\\\nonumber
& = \frac{1}{|Y_{i}|} \frac{|Y_{i}|\prod_{l \in Y_{i}}\exp(x_{l}^{i}) + \sum_{l \in Y_{i}}\exp(x_{l}^{i}) \sum_{k \in \bar{Y_{i}}}\exp(x_{k}^{i})}{\prod_{l \in Y_{i}}(\exp(x_{l}^{i})+  \sum_{k \in \bar{Y_{i}}}\exp(x_{k}^{i})    )} &&  \nonumber
\end{flalign}

\subsection{Bilinear Pooling}
Bilinear based methods~\cite{lin2015bilinear} with deep convolutional neural networks have achieved good results on several fine-grained tasks, such as image recognition~\cite{lin2015bilinear}, and video classification~\cite{ge2016exploiting}. The idea is that in one of the pooling layers, outer-product is performed at each spatial location $(i,j)$ of two networks to generate second-order statistical discriminative local feature representations. The bilinear pooling can be calculated in a pooling layer as follows :
\begin{equation}
p_{i,j} = vec(f^{1}_{i,j},f^{2}_{i,j})
\end{equation}
where $f^a_{i,j}\in \mathbb{R}^{d}$ is a local feature descriptor from one of the pooling layer in network $a$, $p_{i,j}$ is the outer product of two vectors, $vec()$ is the vectorisation operation, and $p\in \mathbb{R}^{d^{2}}$.

\begin{figure}[!t]
\centering
\includegraphics[width=0.9\linewidth]{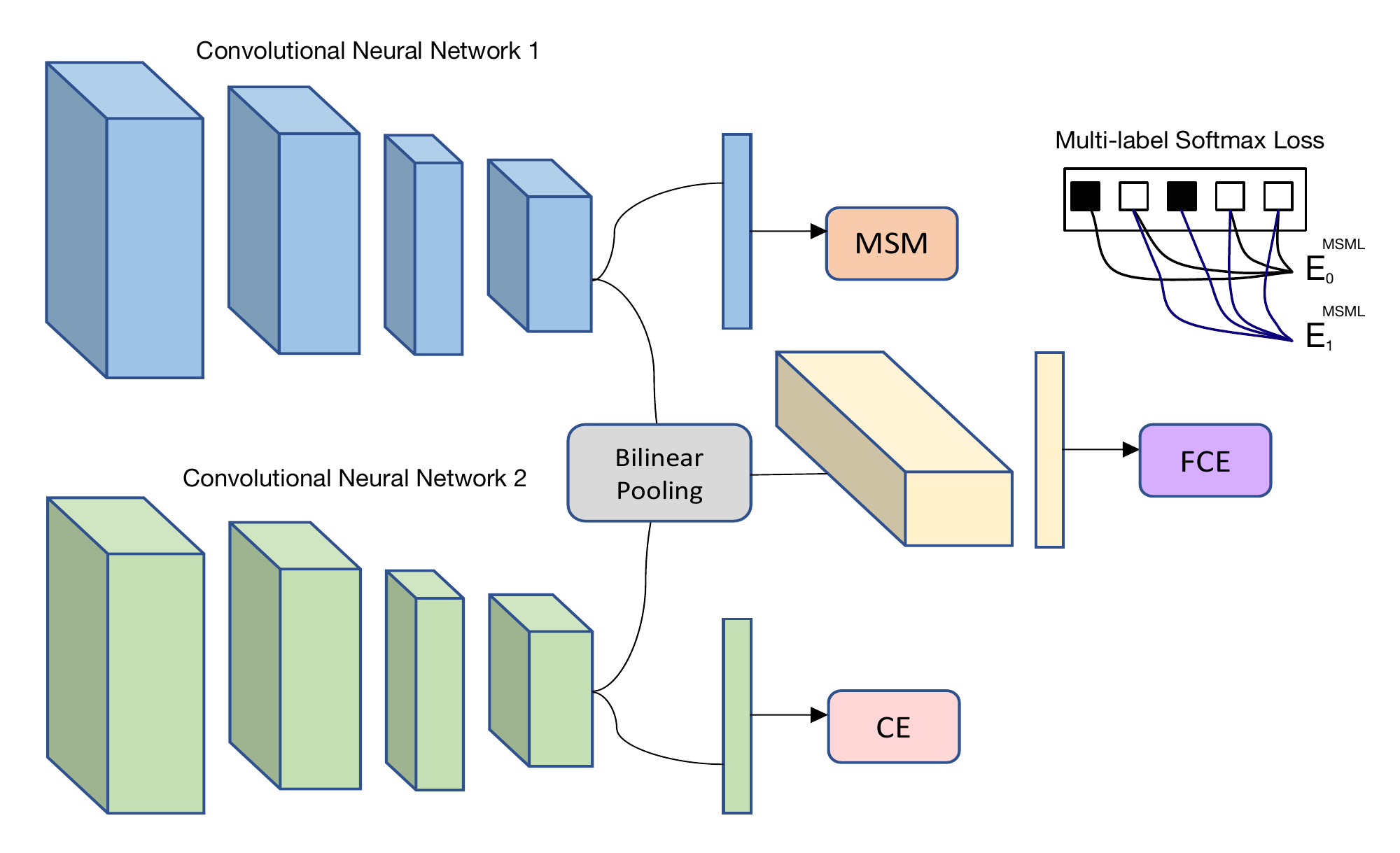}
\caption{Figure shows the architecture of the multi-label fine-grained network for chest x-ray disease detection. MSML loss operates on each positive classes from a sample and encourages them to consider independences between label being present and absent while minimising the oss across classes.}
\label{pic:overview}
\end{figure}

\subsection{Fine-Grained Multi-Label Learning}
\label{FMSML}
As shown in~\cite{lin2015bilinear}, CNNs that are symmetrically initialised with identical parameters may provide efficiency during training. However, the symmetric activations may lead to suboptimal solutions since the model doesn't explore the different space arising from different CNNs. Therefore, in our proposed model we design separate auxiliary losses for two CNN components to break the symmetry between the two feature extractors used for bilinear pooling.

We give an overview of our architecture for fine-grained multi-label learning shown in Fig.\ref{pic:overview}. 
Two separate CNNs are initialised with identical pre-trained parameters. The first CNN stream is attached with normal cross-entropy loss based on sigmoid function. The second CNN's feature maps are input into independent classifiers with MSML loss, which focuses on learning the label correlations. When outputs from bilinear pooling layers are available, fine-grained cross-entropy loss (FCE) can be performed at the fine-grained level.      
Bilinear pooling layer operates on the last convolutional layers of basic architectures, sign sqrt norm and $L2$ norm is applied to the average-pooled bilinear feature layer. A further $1 \times 1$ convolutional layer is added before fed to the classification layer with FCE loss. 
Our model is trained using image-level annotation only for all classes.
The fine-grained multi-label loss to optimiser is the weighted sum of these two:
\begin{equation}
E^{total} = \alpha(E^{MSML} + E^{CE}) + \beta E^{FCE}
\end{equation}
\section{Experiments}


\noindent\textbf{Dataset:} To validate the effectiveness of our methods, we used the ChestX-ray14 dataset introduced by NIH~\cite{wang2017chestx}. This dataset contains 112,120 frontal-view x-rays with 14 diseases labels (multi-labels for each image), which are obtained from associated radiology reports.
We follow the same settings as in~\cite{wang2017chestx} where the entire dataset is randomly split to three sets: 70\% for training, 10\% for validation and 20\% for testing. The split is done at the patient level so tha there is no overlap in different data folds.   

\noindent\textbf{Network and training:}
We use various CNN architectures (ResNet, DenseNet, VGG) with pre-trained parameters on the 2012 ImageNet Challenge as the base CNN component for our evaluated frameworks. 
The ADAM~\cite{kingma2014adam} optimiser with initial learning rate of $10^{-4}$ and $betas=(0.9,0.999)$ is used to optimise the networks. Learning rate decreases by 10 times in every third epochs. 
We use $\alpha=0.2$ and $\beta=0.6$ to weight the proposed multi-label fine-grained network.
Training images are random crops with size $224$ by $224$ from $256$ by $256$ rescaled original images. Input images' value is rearranged to $0-1$ with mean and std norm. All the reported metrics are computed on the test set. 

\noindent\textbf{Evaluation Metrics:}
Two metrics are being used to evaluate the performance of various frameworks. \textbf{Area under the ROC curves (AUC)}: This is the metric used in a few lung x-rays classification literatures~\cite{wang2017chestx,rajpurkar2017chexnet}. The area curve is calculated with Sensitivity ($\frac{TP}{TP+FN}$) as vertical axis and Specificity ($\frac{TN}{TN+FP}$) as horizontal axis. 
There are strong class-imbalance issue for some classes such as Hernia which posses about 2\% of the whole dataset. To track the performance of the majority, we propose to use the \textbf{Weighted Area under the ROC curves (W-AUC)}. Each class AUC is associated with a weighting factor $w_{c}$ with respect to its proportions across the dataset. To drive an insight of how the model performs against normality and abnormality, we further investigate the AUC of disease vs. disease (\textbf{D-AUC}) and disease vs. non-disease (\textbf{N-AUC}) for the test set.     

\subsection{Quantitative Results}
We use the following acronyms. \textbf{R18-CE:} ResNet18 trained with cross entropy (CE) loss, \textbf{R18-BL-CE:} ResNet18 with self bilinear pooling with CE loss, \textbf{R18-F-MSML:} The proposed fine-grained multi-label architecture as described in Sec.~\ref{FMSML}. CNN components use ResNet18 as base architecture. Similar acronyms are denoted for \textbf{D121:} DenseNet-121.     

Using the method proposed on the ChestX-ray14 dataset, we examine the class average AUC and W-AUC. The main results are summarised in Table.~\ref{table:results}. In most instances, we observe that the proposed fine-grained multi-label architecture provides consistent performance improvements for fusion results with the baseline network. The proposed method surpasses the performance of fusion with R18-BL or D121-BL which demonstrates that MSML provides more independences for two CNN components in bilinear-based model.   

Somewhat surprisingly, ensemble model of R18 and D121 doesn't provide higher AUC (first rows under Ensemble Analysis), which indicates that the high-level abstract representation learned from those two models may have strong correlations. This is considered to occur due to baseline results of R18 and D121 offer insignificant performance difference (0.8239 vs. 0.8354) with respect to their model complexity (18 layers vs. 121 layers). The results shown in the last two rows indicate that we can further improve the performance with a base architecture which works better on bilinear pooling function~\cite{ge2017exploiting,lin2017improved}. 


\begin{table}[!t]
\centering 
\caption{\small{AUC and W-AUC results on 14 abnormalities from ChestX-ray14 dataset}}
	\vspace*{0.2cm}
	\begin{tabularx}{1\columnwidth}{l P{2cm} P{1.8cm} P{1.8cm} P{1.8cm} P{1.8cm}}  
	\hline
	{\bf Methods (avg)} &  {\bf AUC} &  {\bf W-AUC} & {\bf D-AUC} & {\bf N-AUC} \\
	\hline
	\bf{Residual Network }  	      &  &  &  & \\	
	\hline
	R18  	      &  0.8239 & 0.5800 & 0.7801 & 0.8552 \\	
	R18-BL 		  &  0.7680  & 0.5543 & 0.6825  & 0.8307 \\
	R18 + R18-BL   	  &  0.8254  & 0.5811 & 0.7756 & 0.8613 \\		
	R18 + R18-F-MSML    & \textbf{0.8388} & \textbf{0.5892} & \textbf{0.7907} & \textbf{0.8733}\\	
	\hline	
	\bf{Dense Network}  	      &  & &  & \\		
	\hline	
	D121  	      &  0.8354 &  0.5888 & 0.7937 & 0.8651 \\	
	D121-BL 		  &  0.8107  &  0.5770 & 0.7523 & 0.8529 \\
	D121 + D121-BL   & 0.8364	  &  0.5896  &  0.7926 & 0.8679  \\		
	D121 + D121-F-MSML    &  \textbf{0.8462} & \textbf{0.5950} & \textbf{0.8011} & \textbf{0.8785}\\	
	\hline	
	\bf{Ensemble Analysis}  	      &  &  &  & \\
	\hline
	R18 + D121	&  0.8355  & 0.5883 & 0.7930 & 0.8653 \\		
	R18  + VGG--F-MSML  	  &  0.8438  &  0.5932 & 0.8011 & 0.8743\\		
	D121 + VGG--F-MSML 	  &  \textbf{0.8537}  & \textbf{0.5952} & \textbf{0.8060} & \textbf{0.8827} \\		
	\hline
	\end{tabularx}
\label{table:results}
\end{table}

\subsection{Training Strategy Analysis}
In this part we examine three training strategies for the proposed method.\footnote{To facilitate the verification process, we use ResNet18 for this experiment.} \textbf{Local:} Train two CNN components with CE and MSML loss then fine-tuning with FCE with bilinear pooling. \textbf{Local-Fixed:} Same as \textbf{Local} but parameters of two CNN components are fixed during final stage fine-tuning with FCE. \textbf{Global:} All parameters are trained simultaneously. We present the performance of those training strategies in column three. Overall, \textbf{Global} makes the best performance of all three. We may infer that global training strategy is beneficial to multi-loss training.

\begin{table}[!t]
  \centering
  \caption{\small{Training strategy of proposed architecture on ChestX-ray14 dataset}}
  \vspace*{0.2cm}  
  \begin{tabular}{|P{2.5cm}|P{2.5cm}|P{2.5cm}|}
    \hline
    \bf Training Strategy & \bf Methods   & \bf AUC    \\ \hline
    Local                 & R18-F-MSML  & 0.8310   \\ \hline
    Local-Fixed           & R18-F-MSML  & 0.8356   \\ \hline
    Global                & R18-F-MSML  & \textbf{0.8388}   \\ \hline
  \end{tabular}
  \label{table:training}
\end{table}

\subsection{Comparable Study to Other Methods}
In this section we show the proposed method can achieve state-of-the-art performance for chest disease detection. Recent work in~\cite{rajpurkar2017chexnet} provided state-of-the-art performance on the ChestX-ray14 dataset. This was achieved by fine-tuning the DenseNet-121 on the dataset. Our work shows that if a fine-grained based  method is used along with a baseline DenseNet, it can achieve better results. Moreover, our method with a much smaller architecture (R-18) can achieve approcimately the same performance as the one using D121 (0.8438 VS. 0.8413). Yao et al.~\cite{yao2017learning} considers label dependency with recurrent model. Both of us used similar label information, but our method is able to achieve better performance with a more flexible model architecture by adding a new loss to explicitly exploiting label correlations rather than embedding the label information into a recurrent model.   

\begin{table}
  \centering
  \caption{\small{Results comparison to other methods from ChestX-ray14 dataset}}
  \begin{tabular}{|P{2.5cm}|P{2.5cm}|P{2.5cm}|}
    \hline
    \bf Methods  & \bf CNN architecture   & \bf AUC    \\ \hline
    Wang et al.\cite{wang2017chestx}                 & R-50 & 0.7363   \\ \hline
    Yao et al.\cite{yao2017learning}          & D121-LSTM   &  0.7980   \\ \hline
    Rajpurkar et al.\cite{rajpurkar2017chexnet}    & D121  &  0.8413   \\ \hline
    Proposed            & R-18/D-121 & \textbf{0.8438}/\textbf{8.8537}   \\ \hline
  \end{tabular}
  \label{table:compare}
\end{table}
\vspace*{-1cm}

\section{Conclusion}
\vspace*{-0.2cm}

In this article, we demonstrate the effectiveness of using multi-label loss function (MSML) learning of DCNN for multi-label lung disease classification on x-ray inputs. We propose two key contributions :(i) the use of fine-grained classification method that learn discriminative representations and (ii) we proposed a novel MSML for deep learning based model which helps to leverage the class dependencies. 
MSML can be interpreted as decomposing the multi-label learning problem into a number of independent classification problems while learning separate distributions for the presence of each class with respect to all the absent classes. The rooted softmax property inside MSML is essential to facilitate the learning process in an exponential-sized output space.  This property makes it appealing to be used as a for multi-label space learning, to predict multiple labels, also making the model alleviate the over-fitting of negative classes because absence classes' outputs are suppressed with respect to each presence class. In medical data, presence of multiple data and imbalance of data is very common. Proposed MSML can handle both of these problems explicitely and embeds the capability into the network. We have illustrated the effectiveness of such approach though an improvement of AUC-ROC score in disease classification in the ChestX-ray14 dataset. However, the similar problem occurs in other medical data from real world. Therefore, the proposed loss function provides a new direction to attain improved performance for wider medical data.



\vspace*{-0.2cm}

\bibliographystyle{IEEEbib}
\bibliography{refs}

\end{document}